\documentclass[conference]{IEEEtran}
\IEEEoverridecommandlockouts
\usepackage{cite}
\usepackage{amsmath,amssymb,amsfonts}
\usepackage{algorithmic}
\usepackage{graphicx}
\usepackage{textcomp}
\usepackage{xcolor}
\def\BibTeX{{\rm B\kern-.05em{\sc i\kern-.025em b}\kern-.08em
    T\kern-.1667em\lower.7ex\hbox{E}\kern-.125emX}}
\begin{document}

\title{Evidence-Grounded SLM Behavioral Prediction Architecture for Future Assistance Use}


\author{
\IEEEauthorblockN{1\textsuperscript{st} Adeyemi Adeseye}
\IEEEauthorblockA{\textit{Department of Computing} \\
\textit{University of Turku}\\
Turku, Finland \\
0009-0006-4184-3436}
\and
\IEEEauthorblockN{2\textsuperscript{nd} Aisvarya Adeseye}
\IEEEauthorblockA{\textit{Department of Computing} \\
\textit{University of Turku}\\
Turku, Finland \\
0009-0003-2401-3076}
\and
\IEEEauthorblockN{3\textsuperscript{rd} Hannu Tenhunen}
\IEEEauthorblockA{\textit{Electronic System Design} \\
\textit{Royal Institute of Technology}\\
Stockholm, Sweden \\
0000-0003-1959-6513}
\and
\IEEEauthorblockN{4\textsuperscript{th} Jouni Isoaho}
\IEEEauthorblockA{\textit{Department of Computing} \\
\textit{University of Turku}\\
Turku, Finland \\
0000-0002-5789-3992}
\vspace{-0.8cm}
}

\maketitle

\begin{abstract}

Users often exhibit changes in interaction behavior before explicitly seeking assistance, enabling systems to anticipate future assistance use. This is especially important in high-stakes environments, where delayed assistance can have operational, financial, or safety consequences. However, conventional Large Language Models (LLMs) are often unsuitable for continuous on-device behavioral prediction because of their computational and memory demands. Behavioral patterns are also highly individual and device-specific, motivating local processing that preserves privacy and data sovereignty. Small Language Models (SLMs) offer a promising alternative, but their ability to reason reliably over low-level sequential interaction events remains underexplored. This study introduces Evidence-Grounded Local Behavioral Prediction (ELBP), a four-layer architecture combining deterministic interaction processing, dual-view behavioral representation, SLM-based prediction, and evidence verification with abstention. Three instruction-tuned models of 3B parameters or fewer were evaluated using raw, structured, dual-view, and complete ELBP configurations. The best-performing model, Qwen2.5-3B with ELBP, achieved a full-set F1-score of 0.80 and PR-AUC of 0.84, representing a 15.9\% relative improvement over raw-event prompting and 17.6\% over XGBoost, while full-set ECE decreased by 43.2\% relative to raw-event prompting. With evidence verification and abstention, selective F1 increased to 0.84 at 91\% coverage and evidence support reached 96\%. These results show that structured behavioral representation and evidence-verified selective prediction can improve the reliability of behavioral forecasting in SLMs.

\end{abstract}

\begin{IEEEkeywords}
local large language models, behavioral prediction, help-seeking prediction, interaction logs, evidence grounding, behavioral representation, adaptive systems
\end{IEEEkeywords}

\section{Introduction}
\label{sec:introduction}

Users often change their interaction behavior before explicitly requesting assistance. These changes may precede subsequent assistance-seeking behavior, allowing systems to anticipate future assistance use. Prior educational data-mining studies show that students' initial problem-solving attempts and patterns of hint use can predict subsequent assistance-seeking behavior~\cite{a1}. Large-scale datasets such as EdNet further demonstrate the value of sequential interaction traces for examining how earlier actions relate to later help-seeking events~\cite{a2}. This capability is particularly important in high-stakes environments, where delayed assistance may have operational, financial, or safety consequences.
Recent research has explored Large Language Models (LLMs) for modeling user behavior from previous interaction histories~\cite{a3}. However, conventional LLMs are often impractical for continuous on-device behavioral prediction because of their computational and memory requirements~\cite{a5,a6,a7}. Behavioral patterns are also highly individual and device-specific, creating a need for local processing that can maintain device-level behavioral profiles while supporting privacy and data sovereignty~\cite{a8}. 
Small Language Models (SLMs) therefore provide a promising alternative for local behavioral prediction. However, their ability to reason reliably over long, low-level sequential interaction events remains underexplored. Prediction quality can also depend strongly on how interaction histories are represented~\cite{a4}. Structured behavioral representations and verification mechanisms may therefore help smaller models identify relevant patterns, ground predictions in observed evidence, and abstain when evidence is insufficient~\cite{a9,a10}. 

This study therefore aims to develop and evaluate an \textit{Evidence-Grounded Local Behavioral Prediction} (ELBP) architecture for predicting future assistance use from sequential interaction traces using SLMs.

The study addresses the following research questions:

\begin{itemize}
    \item \textbf{RQ1:} How accurately can local SLMs predict future assistance use from recent interaction histories?
    
    \item \textbf{RQ2:} To what extent do structured behavioral representation and evidence verification improve prediction reliability over direct reasoning over raw interaction logs for SLMs?
\end{itemize}

The contribution of this study is a four-layer local SLM architecture that combines deterministic behavioral processing, dual-view interaction representation, evidence-grounded prediction, and post-prediction verification with abstention to improve SLM reasoning over raw interaction logs.

\section{Related Work}
\label{sec:related_work}

Prior assistance prediction studies have mainly relied on conventional models and engineered behavioral features. Beck \textit{et al.} predicted learners' help requests in an intelligent reading tutor~\cite{c1}, while Castro \textit{et al.} used previous attempts and hint histories to predict future use ~\cite{a1}. 
Recently, LLMs have been used for behavioral modeling. Nguyen \textit{et al.} showed that LLMs can use observed interaction histories to model subsequent student behavior~\cite{a3}. Ober \textit{et al.} further showed that LLMs could be used to identify assistance use and persistence indicators from learner dialogues~\cite{c4}. However, these studies do not examine local SLMs for short-horizon prediction from low-level interaction events.
Representation is equally important. Krishnamurthy \textit{et al.} showed that summarized interaction histories can substantially affect LLM decision behavior~\cite{a4}, while structured semantic processing and validation have improved reliability in local LLM pipelines~\cite{a9}. Edge and on-device LLM research further identifies computation, memory, latency, and privacy as key deployment concerns~\cite{a5,a6,a7}, with local processing offering additional privacy benefits for sensitive data~\cite{a8}. Verification is also important because language-model predictions may be weakly supported by evidence; structured validation can reduce unsupported outputs~\cite{a9,a10}.
To solve this, ELBP integrates these directions through local SLMs, structured behavioral representations, evidence grounding, and abstention to predict near-term assistance-seeking behavior.

\section{Evidence-Grounded Local Behavioral Prediction Architecture}
\label{sec:architecture}
\begin{figure*}[!t]
    \centering
    \includegraphics[width=0.75\textwidth]{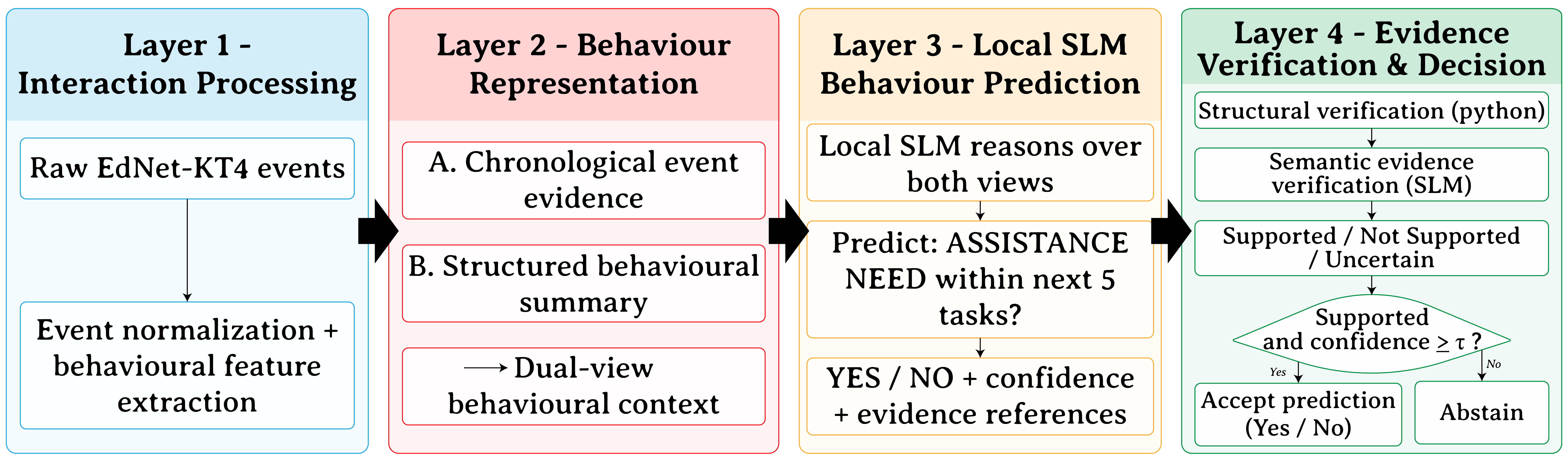}
    \caption{Evidence-Grounded Local Behavioral Prediction Architecture for Assistance Use}
    \label{fig:elbp_architecture}
\end{figure*}

This study proposes \textit{Evidence-Grounded Local Behavioral Prediction} (ELBP), a four-layer architecture for forecasting near-term user behavior from sequential interaction traces using locally deployed SLMs. The design is primarily motivated by SLMs' reasoning capacity and limited context budgets. As illustrated in Fig.~\ref{fig:elbp_architecture}, the architecture separates deterministic interaction processing and behavioral representation from SLM-based prediction and evidence verification.

This separation is intentional: low-level operations such as event ordering, temporal calculations, aggregation, and feature construction are performed deterministically before SLM inference. The local SLM is therefore used primarily for interpreting behavioral patterns and forecasting the target behavior, rather than for data cleaning or arithmetic operations. The final layer evaluates whether a prediction is supported by the behavioral evidence supplied to the model and permits abstention when sufficient support is unavailable.
The architecture is designed around three principles: \textit{local inference}, \textit{evidence-grounded prediction}, and \textit{selective decision-making}. Local inference enables behavioral traces to remain within local or organizational infrastructure and reduces dependence on cloud-based services. Evidence grounding makes the basis of each prediction inspectable, while selective decision-making prevents the system from being forced to issue a prediction when the available behavioral evidence is insufficient or contradictory.

%
%

\subsection{Layer 1: Interaction Processing}

The first layer converts raw interaction traces into consistent and interpretable behavioral observations. Raw event logs typically represent individual system actions rather than complete behavioral episodes. Consequently, directly providing such logs to an SLM would require the model to simultaneously perform event grouping, temporal arithmetic, sequence interpretation, and behavioral reasoning, imposing substantial reasoning demands on the SLM. The objective of Layer~1 is therefore to transform low-level interaction events into reliable behavioral observations while preserving their temporal ordering and provenance. The processing includes chronological ordering, grouping of related events, calculation of temporal intervals, identification of repeated or revised actions, normalization of event labels, treatment of missing values, and derivation of task-level outcomes.

For example, a raw sequence containing an activity entry, multiple responses, and a submission can be converted into an interaction-level record containing the final response, number of revisions, elapsed response time, and observed outcome. The transformation is deterministic and does not involve SLM inference.
Layer~1 is implemented in Python. Libraries such as \texttt{pandas}, \texttt{NumPy}, and standard JSON-processing utilities were utilized for event reconstruction and behavioral feature extraction. \texttt{PyArrow} was used for large event collections and column-oriented processing.

%

\subsection{Layer 2: Behavior Representation}

\begin{figure*}[!t]
    \centering
    \includegraphics[width=0.75\textwidth]{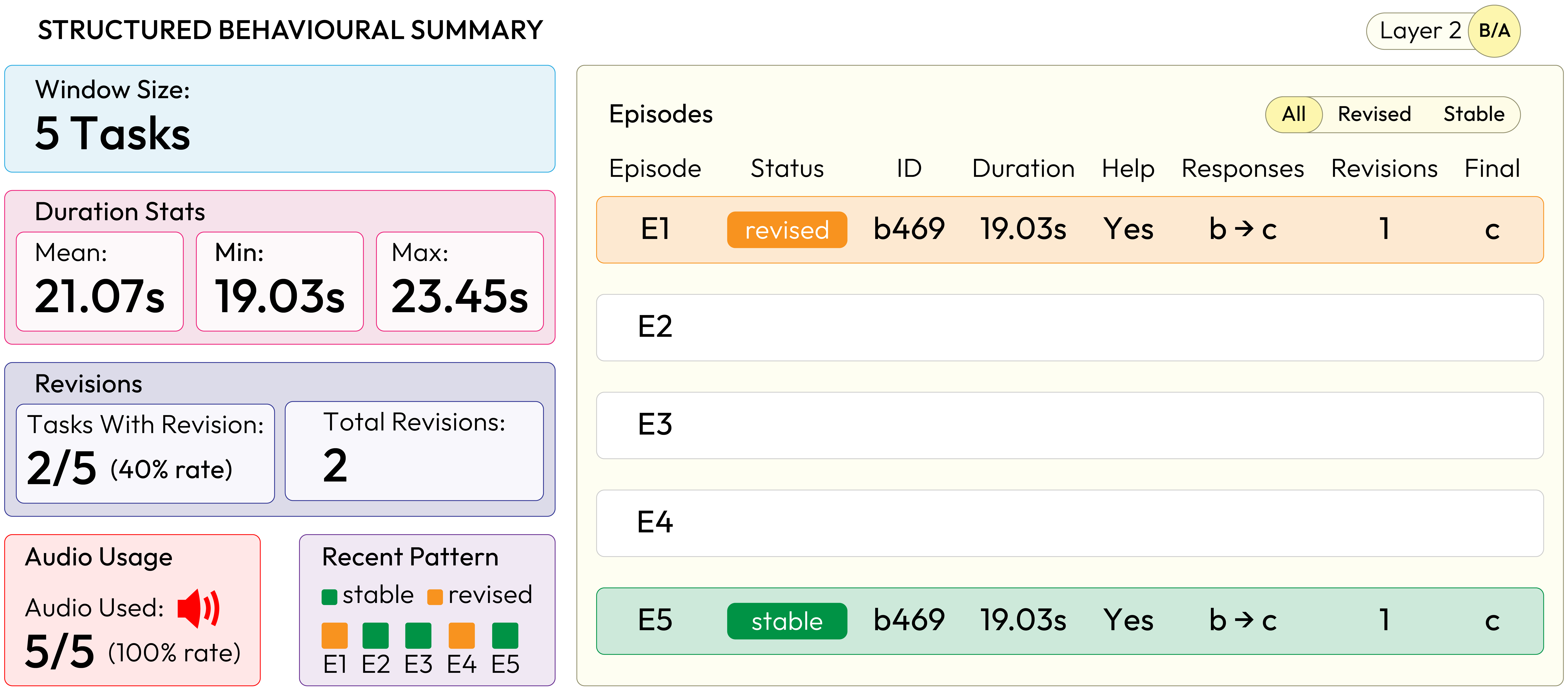}
    \caption{Graphical view of the textual representation of information that is passed from Layer 2 to Layer 3}
    \label{fig:layer 2}
\end{figure*}

As seen in Fig.~\ref{fig:layer 2}, Layer~2 converts the processed interaction history into a dual-view representation suitable for local-SLM reasoning. The central design assumption is that chronological event sequences and structured behavioral summaries provide complementary information. The chronological representation preserves \textit{when} and \textit{in what order} behavioral events occurred, whereas the structured representation explicitly exposes higher-level patterns that an SLM would otherwise need to calculate internally.
For user $u$ at prediction point $t$, Layer~2 produces

{\small
\begin{equation}
R_{u,t} = \left(E_{u,t}, S_{u,t}\right),
\label{eq:dual_view}
\end{equation}
}
where $E_{u,t}$ denotes the chronological evidence representation and $S_{u,t}$ denotes the structured behavioral summary.

\subsubsection{Chronological Evidence Representation}

The chronological view preserves the temporal sequence of observations. Each observation is assigned a unique evidence identifier, allowing later SLM predictions to refer explicitly to the events used as justification. A sequence may represent increasing response duration, repeated response revisions, or changes to assistance use. The event-level representation enables the SLM to reason about trajectories and transitions that may disappear when behavior is represented only by aggregate statistics. Such patterns include repeated failures, increasing hesitation, recovery following difficulty, and transitions from stable to unstable interaction behavior.

\subsubsection{Structured Behavioral Summary}

The second view summarizes the same interaction window using deterministic behavioral variables. Candidate summary variables include recent performance rate, consecutive unsuccessful outcomes, mean response duration, response-duration trend, number of response revisions, number of retries or reversals, and previous assistance-seeking behavior.
The summary provides an explicit abstraction of the event sequence and reduces the amount of low-level computation required. This is particularly relevant for an SLM because of its limited reasoning capacity and context budgets.


%
%

\subsection{Layer 3: Local SLM Behavior Prediction}

Layer~3 constitutes the main reasoning component of ELBP. A locally deployed SLM receives the dual-view representation generated by Layer~2 and forecasts the predefined near-term behavioral outcome.
The model receives both the chronological evidence and the structured behavioral summary and is instructed to reason only from the supplied information. Rather than returning an unexplained class label, the model is required to output the predicted behavioral class, a confidence value, the identifiers of the evidence supporting the prediction, and a concise justification.
Formally,
{\small
\begin{equation}
f_{\theta}(R_{u,t})
\rightarrow
\left(
\hat{y}_{u,t},
c_{u,t},
Z_{u,t},
r_{u,t}
\right),
\label{eq:prediction}
\end{equation}
}
where $\hat{y}_{u,t}$ denotes the predicted behavior, $c_{u,t}$ is the model-reported confidence, $Z_{u,t}$ contains the cited evidence identifiers, and $r_{u,t}$ represents the behavioral justification.
A structured output takes the following form:

{\scriptsize
\begin{verbatim}
{
  "prediction": "YES",
  "confidence": 0.84,
  "evidence_ids": ["E3", "E5", "E8"],
  "reason": "Recent failures are
  accompanied by increasing response
  duration and repeated response
  revisions."
}
\end{verbatim}
}
Requiring evidence identifiers makes the prediction auditable and establishes an explicit connection between the model output and the supplied behavioral trace.
The prediction layer is implemented via a fixed \textbf{prompt} executed on a local SLM. The inference framework uses \texttt{llama.cpp}. However, other inference frameworks such as Ollama, Hugging Face Transformers, or vLLM may be used depending on the selected model and hardware. The base architecture does not require model fine-tuning, allowing multiple local SLMs to be evaluated under identical prompting conditions.

%
%

\subsection{Layer 4: Evidence Verification and Decision}

Layer~4 separates prediction generation from prediction acceptance. Although an SLM may produce a plausible behavioral forecast, its justification can contain unsupported, exaggerated, or incorrectly referenced evidence. This layer therefore verifies the grounding of the prediction before issuing a final decision.
The verification stage combines deterministic Python checks with a focused SLM verification prompt.

\subsubsection{Structural Verification}

Python first validates the structure of the Layer~3 output. The checks include JSON-schema compliance, validity of the prediction label, confidence-range validation, presence of required evidence identifiers, and confirmation that every cited evidence identifier exists in the supplied interaction history. This stage detects malformed outputs and fabricated evidence references without requiring an additional SLM call.

\subsubsection{Semantic Evidence Verification}

Deterministic checks validate output structure and confirm that cited evidence identifiers exist in the supplied interaction history, but they do not assess whether that evidence supports the model’s behavioral interpretation. A focused SLM verifier therefore evaluates whether the justification is semantically supported by the cited evidence and classifies the relationship as

\begin{equation}
v_{u,t}
\in
\{
\text{SUPPORTED},
\text{NOT\_SUPPORTED},
\text{UNCERTAIN}
\}.
\label{eq:verification}
\end{equation}

The verifier does not generate a new prediction; it only assesses whether the model's reasoning is supported by the supplied evidence.

\subsubsection{Selective Final Decision}

The verification result and confidence value are subsequently combined using deterministic decision rules. The final output is
{\small
\begin{equation}
D_{u,t} =
\begin{cases}
\hat{y}_{u,t},
&
v_{u,t}=\text{SUPPORTED}
\land
c_{u,t}\geq\tau,\\
\text{ABSTAIN},
&
\text{otherwise},
\end{cases}
\label{eq:abstention}
\end{equation}
}
where $\hat{y}_{u,t}$ is the predicted class, $c_{u,t}$ is model confidence, and $\tau$ is the confidence threshold. The threshold $\tau$ is selected on the development set to maximize selective F1 while maintaining at least 90\% coverage, and is then fixed for the held-out test set. Abstention withholds predictions with insufficient confidence or evidence support, enabling performance to be evaluated jointly with the proportion of cases receiving a decision (coverage).

\section{Experimental Setup, Benchmarks, and Evaluation}
\label{sec:experimental_setup}

\subsection{Testbed and Gold Standard}

The experimental testbed uses EdNet-KT4, the most fine-grained level of
the EdNet dataset, which contains sequential records of learner interactions
rather than only final question responses~\cite{a2}. The complete KT4
dataset contains 131,441,538 interaction records from 297,915 learners.
For computational tractability, 70\% of KT4 learners were randomly sampled using seed 42, with all interaction records for the selected learners retained to preserve complete sequential histories. This scale provides a large sequential testbed while maintaining the fine-grained interaction structure required for predicting near-term behavioral events.
The extracted KT4 fields such as \texttt{timestamp}, \texttt{action\_type},
\texttt{item\_id},
\texttt{user\_answer}, and question metadata \texttt{correct\_answer} were required to derive correctness, response duration, answer revisions, choice reversals, recent correctness, and previous assistance use deterministically. Events unrelated to learning behavior are excluded.

EdNet does not provide a predefined label for near-term assistance use. An operational behavioral gold standard is therefore derived deterministically from observed future actions. Prediction points are defined at the end of each task. At each prediction point, the model receives interaction information from the preceding five tasks. A positive label is assigned when the learner voluntarily requests assistance, i.e., checks an explanation or lecture, within the subsequent five tasks; otherwise, the instance is labeled negative. System-initiated assistance, such as \texttt{adaptive\_offer}, is excluded. Data are partitioned by learner into training, development, and test sets to prevent histories from the same learner appearing across partitions. Learners were partitioned into training, development, and test sets using a 70/15/15 learner-disjoint split.

\subsection{Local SLM Configuration}

Three instruction-tuned models representing different model families and increasing small-model capacities are evaluated: \textbf{Llama-3.2-1B-Instruct}, \textbf{Gemma-2-2B-IT}, and \textbf{Qwen2.5-3B-Instruct}~\cite{b6,b7,b8}. All models are executed locally using the same inference engine and comparable 4-bit quantization on a 64 GB RAM and 8GB VRAM local infrastructure. No fine-tuning is performed. The same zero-shot task instruction, output schema, decoding parameters, and experimental instances are used across models. Greedy decoding was used to reduce sampling variability and improve reproducibility across model and architectural comparisons.

\subsection{Benchmark Configurations}

Four configurations isolate the contribution of each ELBP component.
\textbf{B0--Raw} receives the original chronological KT4 events without Layer-1 feature construction directly to the local SLM. \textbf{B1--Structured} supplies only the deterministic
behavioral summary. \textbf{B2--Dual-View} provides both chronological
evidence and the structured summary. \textbf{B3--ELBP} adds evidence
attribution, verification, and abstention to B2. B0--B2 therefore provide
controlled architectural baselines rather than different prediction
algorithms.
Additionally, \textbf{XGBoost} is included as a non-SLM baseline
to determine whether the local SLM provides predictive value beyond the
deterministic behavioral variables alone. XGBoost receives the same
structured behavioral variables supplied to B1, including recent
performance rate, consecutive unsuccessful outcomes, response-duration
statistics, answer revisions, retries or reversals, and previous assistance-seeking behavior. Model fitting and hyperparameter selection
are performed only on the training/development data, while the same
learner-disjoint test partition used for the SLM configurations is retained
for final evaluation. This provides a conventional supervised-learning
reference under the same prediction target and input information
conditions~\cite{b12}.


\subsection{Evaluation Metrics}

Classification performance is measured using \textbf{Precision}, \textbf{Recall}, and \textbf{F1-Score}:
{\small
\begin{equation}
\mathrm{Precision}=\frac{TP}{TP+FP},\qquad
\mathrm{Recall}=\frac{TP}{TP+FN},
\end{equation}
}
{\small
\begin{equation}
\mathrm{F1}=2\frac{\mathrm{Precision}\cdot\mathrm{Recall}}
{\mathrm{Precision}+\mathrm{Recall}},
\end{equation}
}
where $TP$, $FP$, and $FN$ denote true positives, false positives, and false negatives respectively. \textbf{PR-AUC} measures performance across classification thresholds and is suitable for imbalanced positive events~\cite{b9}.



Calibration is measured using \textbf{Expected Calibration Error (ECE)}~\cite{b11},
{\small
\begin{equation}
\mathrm{ECE}=\sum_{m=1}^{M}\frac{|B_m|}{N}
\left|\mathrm{acc}(B_m)-\mathrm{conf}(B_m)\right|,
\end{equation}
}

where $B_m$ is confidence bin $m$, and $\mathrm{acc}(B_m)$ and $\mathrm{conf}(B_m)$ are its empirical accuracy and mean confidence.

ELBP reliability is measured using \textbf{Evidence Support} and \textbf{Coverage}. 
{\small
\begin{equation}
\mathrm{EvidenceSupport}=
\frac{N_{S}}
{N_{S}+N_{NS}+N_{U}},
\end{equation}
}

where $N_{S}$, $N_{NS}$, and $N_{U}$ denote the numbers of
SUPPORTED, NOT\_SUPPORTED, and UNCERTAIN verification outcomes, respectively.
{\small
\begin{equation}
\mathrm{Coverage}=
\frac{N_{\mathrm{YES}}+N_{\mathrm{NO}}}
{N_{\mathrm{YES}}+N_{\mathrm{NO}}+N_{\mathrm{ABSTAIN}}}.
\end{equation}
}
where $N_{\mathrm{YES}}$ and $N_{\mathrm{NO}}$ denote accepted final predictions,
while $N_{\mathrm{ABSTAIN}}$ denotes predictions withheld by the selective decision mechanism.
\textbf{Selective F1} is calculated only on non-abstained predictions. \textbf{Local Deployment Efficiency} is reported using inference latency, throughput (tokens/s), and peak VRAM usage.

\section{Results}
\label{sec:results}

\subsection{Overall Prediction Performance}
Table~\ref{tab:overall_results} compares the majority-class reference baseline, the non-SLM XGBoost baseline, and the three local SLMs under
the direct raw-event condition (B0) and the ELBP prediction configuration prior to selective acceptance. To ensure a fair comparison across
configurations, the precision, recall, F1, PR-AUC, and ECE values reported in this table are calculated over the complete test set without selective post-abstention. Performance with selective post-abstention is reported separately in Section~\ref{sec:architectural_benchmark}.
The majority baseline always predicts no future assistance use and therefore obtains 0 recall and F1 for future assistance-use events. The XGBoost baseline achieves an F1 of 0.68 and PR-AUC of 0.72, providing a stronger non-SLM reference
for evaluating the local SLM configurations.
Direct raw-event reasoning produces performance close to the conventional XGBoost machine learning baseline. Llama-3.2-1B achieves an F1 of 0.62, Gemma-2-2B reaches 0.67, and Qwen2.5-3B reaches 0.69, slightly exceeding XGBoost. Applying the ELBP representation and prediction architecture improves full-set performance for all three models. Llama-3.2-1B increases from 0.62 to 0.70, Gemma-2-2B from 0.67 to 0.75, and Qwen2.5-3B from 0.69 to 0.80.
Qwen2.5-3B with ELBP achieves the highest full-set precision (0.81), recall (0.79), F1 (0.80), and PR-AUC (0.84). Its pre-selection ECE decreases from 0.118 under B0 to 0.067 after dual-view representation. Following evidence verification and abstention, ECE decreases to 0.051 on retained predictions. Compared to XGBoost, the complete Qwen2.5-3B ELBP configuration improves
the full-set F1 by 0.12.

\begin{table}[t]
\centering
\caption{Full-Set Prediction Performance Across Baselines and Local SLMs}
\label{tab:overall_results}
\resizebox{\columnwidth}{!}{
\begin{tabular}{llcccccc}
\hline
\textbf{Model} &
\textbf{Configuration} &
\textbf{Prec.} &
\textbf{Recall} &
\textbf{F1} &
\textbf{PR-AUC} &
\textbf{Full ECE}$\downarrow$  \\
\hline

Majority class & Always-NO
& -- & 0.00 & 0.00 & 0.24 & --  \\

XGBoost & Structured features
& 0.70 & 0.66 & 0.68 & 0.72 & 0.084  \\

Llama-3.2-1B & B0 Raw
& 0.65 & 0.59 & 0.62 & 0.67 & 0.142 \\

Llama-3.2-1B & ELBP (pre-selection)
& 0.72 & 0.69 & 0.70 & 0.75 & 0.099  \\

Gemma-2-2B & B0 Raw
& 0.70 & 0.64 & 0.67 & 0.72 & 0.126  \\

Gemma-2-2B & ELBP (pre-selection)
& 0.77 & 0.74 & 0.75 & 0.80 & 0.078  \\

Qwen2.5-3B & B0 Raw
& 0.72 & 0.66 & 0.69 & 0.73 & 0.118  \\

Qwen2.5-3B & ELBP (pre-selection)
& \textbf{0.81} & \textbf{0.79} &
\textbf{0.80} & \textbf{0.84} &
\textbf{0.067}  \\

\hline
\end{tabular}}
\end{table}

\subsection{Architectural Benchmark Comparison}
\label{sec:architectural_benchmark}

To isolate the contribution of behavioral representation, verification, and abstention, four Qwen2.5-3B-Instruct configurations were evaluated alongside XGBoost. Table~\ref{tab:benchmark_results} reports full-set and selective F1 separately to distinguish gains from selective post-abstention.
XGBoost achieves a full-set F1 of 0.68, while B0 raw-event reasoning reaches 0.69. Replacing raw events with the structured behavioral summary (B1) increases F1 to 0.75, and combining the summary with chronological evidence (B2) further increases it to 0.80, indicating that the two representations provide complementary predictive information. B3, after evidence verification and abstention, increases selective F1 to  0.84 at 0.91 coverage, withholding 9\% of cases. Evidence support reached 0.96 in B3, while ECE decreases to 0.051.
Relative to XGBoost, B3 improves full-set F1 by 0.12, or 17.6\%, while the improvement over B0 is 0.11, or 15.9\%. The small difference between XGBoost and B0, together with the improvements from B0 through B2, suggests that much of the observed predictive gain is associated with the ELBP representation architecture. Verification and abstention provide an additional reliability gain, with F1 increasing from 0.80
on the full set to 0.84 on retained predictions while maintaining 91\% coverage.

\begin{table}[t]
\centering
\caption{Qwen2.5-3B-Instruct Architectural and Non-SLM Benchmark Comparison}
\label{tab:benchmark_results}
\resizebox{\columnwidth}{!}{
\begin{tabular}{lccccc}
\hline
\textbf{Configuration} &
\textbf{Full F1} &
\textbf{Sel. F1} &
\textbf{ECE}$\downarrow$ &
\textbf{Evidence Support}$\uparrow$ &
\textbf{Coverage} \\
\hline

XGBoost
& 0.68 & -- & 0.084 & -- & 1.00 \\

B0 Raw
& 0.69 & -- & 0.118 & -- & 1.00 \\

B1 Structured
& 0.75 & -- & 0.083 & -- & 1.00 \\

B2 Dual-View
& \textbf{0.80} & -- & 0.067 & -- & 1.00 \\

B3 ELBP
& \textbf{0.80}
& \textbf{0.84}
& \textbf{0.051 (Sel.)}
& \textbf{0.96}
& 0.91 \\

\hline
\end{tabular}}
\vspace{0.1mm}

\footnotesize{
For B3, ECE is calculated on non-abstained predictions; all other ECE values are calculated on the full test set.
}
\end{table}

\subsection{Ablation Study}

Table~\ref{tab:ablation} reports the ablation analysis of the complete
ELBP architecture using Qwen2.5-3B. Both full-set and selective F1 are
reported because some ablations alter the proportion of cases accepted
by the final decision mechanism. When semantic evidence verification is
ablated, structural validation is retained and the selective decision is
based on the confidence threshold alone.
Removing the structured behavioral summary produces the largest reduction
in performance, decreasing full-set F1 from 0.80 to 0.74 and
selective F1 from 0.84 to 0.77. Although this remained above
the XGBoost baseline of 0.68, the reduction demonstrates that deterministic
behavioral summarization contributes substantially to the performance of ELBP. Removing chronological evidence reduces full-set F1 from 0.80 to 0.77,
and selective F1 from 0.84 to 0.80, indicating that temporal event sequences contribute information that is not fully preserved by aggregate behavioral variables. The two representations therefore provide complementary information within the dual-view architecture. Removing semantic verification increases coverage from 0.91 to 0.93. It also increases ECE from 0.051 to 0.064. When abstention is removed entirely, coverage returns to 1.00, and F1 corresponds to the full-set B3 value of 0.80. The difference between the full-set F1 of 0.80 and
selective F1 of 0.84 therefore quantifies the performance--coverage
trade-off introduced by selective prediction.

\begin{table}[t]
\centering
\caption{Ablation Results for Qwen2.5-3B}
\label{tab:ablation}
\resizebox{\columnwidth}{!}{
\begin{tabular}{lccccc}
\hline
\textbf{Configuration} &
\textbf{Full F1} &
\textbf{Sel. F1} &
\textbf{ECE}$\downarrow$ &
\textbf{Evidence Support}$\uparrow$ &
\textbf{Coverage} \\
\hline

Full ELBP
& 0.80 & \textbf{0.84}
& \textbf{0.051} & \textbf{0.96} & 0.91 \\

-Structured Behavioral Summary
& 0.74 & 0.77
& 0.079 & 0.94 & 0.89 \\

-Chronological Evidence Representation
& 0.77 & 0.80
& 0.064 & -- & 0.92 \\

-Semantic Evidence Verification
& 0.80 & 0.81
& 0.064 & -- & 0.93 \\

-Abstention
& 0.80 & 0.80
& 0.067 & 0.96 & 1.00 \\

\hline
\end{tabular}}
\vspace{0.1mm}

\footnotesize{ECE uses retained predictions under abstention \& full set at full coverage; verifier ablation retains structural checks \& confidence-threshold acceptance.}
\end{table}


\subsection{Local Deployment Efficiency}

Table~\ref{tab:efficiency} summarizes the computational requirements of
the complete ELBP architecture. Reported ELBP latency is measured end-to-end and includes both Layer-3 prediction and Layer-4 semantic verification. Because the final ELBP decision mechanism
permits abstention, performance is reported as selective F1
together with coverage. XGBoost is not included in this table because
token throughput, model quantization, and VRAM requirements are SLM-specific measures.
Llama-3.2-1B provides the lowest computational cost among the ELBP
configurations, achieving a selective F1 of 0.74 at 0.92 coverage, with
a latency of 0.93~s, throughput of 76 tokens/s, and peak VRAM usage of
1.4~GB. Gemma-2-2B achieves a selective F1 of 0.79 at 0.91 coverage
while requiring 1.41~s per inference and 2.1~GB of peak VRAM.
Qwen2.5-3B achieves the highest selective F1 of 0.84 at 0.91 coverage,
with a latency of 1.87~s and peak VRAM consumption of 2.8~GB.

The results show a clear performance--efficiency trade-off. XGBoost provides substantially lower inference cost, with an average inference latency of 0.03 s per instance, but achieves lower predictive performance and does not provide evidence-grounded reasoning or selective verification. Among the local SLM configurations, Llama-3.2-1B provides the lowest-latency operating point, Gemma-2-2B provides an intermediate configuration, and Qwen2.5-3B provides the highest performance at a higher computational cost.

\begin{table}[t]
\centering
\caption{Local Deployment Efficiency of ELBP}
\label{tab:efficiency}
\resizebox{\columnwidth}{!}{
\begin{tabular}{lcccccc}
\hline
\textbf{Model} &
\textbf{Sel. F1} &
\textbf{Coverage} &
\textbf{Latency (s)} &
\textbf{Tokens/s} &
\textbf{Peak VRAM} &
\textbf{Quant.} \\
\hline

Llama-3.2-1B
& 0.74 & \textbf{0.92} & \textbf{0.93} &
\textbf{76} & \textbf{1.4 GB} & Q4 \\

Gemma-2-2B
& 0.79 & 0.91 & 1.41 &
59 & 2.1 GB & Q4 \\

Qwen2.5-3B
& \textbf{0.84} & 0.91 & 1.87 &
47 & 2.8 GB & Q4 \\

\hline
\end{tabular}}
\end{table}

\section{Conclusion}
\label{sec:conclusion}
This study introduced ELBP, a four-layer architecture for predicting future assistance use using SLMs. The proposed design combines deterministic interaction processing, dual-view behavioral representation, evidence-grounded prediction, and post-prediction verification with abstention. The results indicate that structured behavioral representation can improve prediction over raw interaction logs, while verification improves reliability. A key limitation is that the study evaluates a single behavioral domain and prediction target. Future work will therefore test ELBP across additional interaction domains and behavioral outcomes to assess whether the architecture generalizes beyond assistance-seeking prediction.

\end{document}